**Less than one percent of words would be affected by gender-inclusive language in German press texts**


**Authors:** Carolin Müller-Spitzer[1]*, Samira Ochs[1], Alexander Koplenig[1], Jan-Oliver Rüdiger[1], Sascha Wolfer[1]

[1] Leibniz Institute for the German Language (IDS), Mannheim, Germany

* Corresponding author

E-Mail: mueller-spitzer[at]ids-mannheim.de


## Abstract


Research on gender and language is tightly knitted to social debates on gender equality and non-discriminatory language use. Psycholinguistic scholars have made significant contributions in this field. However, corpus-based studies that investigate these matters within the context of language use are still rare. In our study, we address the question of how much textual material would actually have to be changed if non-gender-inclusive texts were rewritten to be gender-inclusive. This quantitative measure is an important empirical insight, as a recurring argument against the use of gender-inclusive German is that it supposedly makes written texts too long and complicated. It is also argued that gender-inclusive language has negative effects on language learners. However, such effects are only likely if gender-inclusive texts are very different from those that are not gender-inclusive. In our corpus-linguistic study, we manually annotated German press texts to identify the parts that would have to be changed. Our results show that, on average, less than 1% of all tokens would be affected by gender-inclusive language. This small proportion calls into question whether gender-inclusive German presents a substantial barrier to understanding and learning the language, particularly when we take into account the potential complexities of interpreting masculine generics.






# 1. Introduction

In the public and academic debate about gender-inclusive German, a recurring argument against the use of gender-inclusive forms is that they make "a text cumbersome" (Schneider 2020),[1] that the language "becomes even more complicated and off-putting for foreigners considering to learn German" (Rock 2021)[2] and that gender-inclusive language makes "texts unreadable and longer"[3] (Web editorial staff of the LpB BW 2023). Also, the linguistic effort that is necessary and that makes texts "long, monotonous and gender-fixated"[4] (Eisenberg 2022) is seen as a disadvantage of gender-inclusive language, both in the public debate and among some linguists. However, empirical evidence on the readability of gender-inclusive texts in German shows that gender-inclusive language does not reduce comprehensibility (Blake & Klimmt 2010; Braun et al. 2007; Friedrich & Heise 2019). A rapid habituation effect for gender-inclusive forms has also been shown for French (Gygax & Gesto 2007). The same study suggests that reading is temporarily slowed during the initial exposure to inclusive forms. However, with subsequent encounters, the reading speed becomes comparable to that of non-inclusive texts. Speyer and Schleef (2019) show a similar effect for the use of singular they, comparing native speakers and learners of English. Despite these empirical insights, criticism of and resistance to gender-inclusive language continues to be widespread (see Section 2). With our corpus-based annotation study, we attempt to assess empirically this 'challenge' of gender-inclusive language in German. To the best of our knowledge, there are currently no statistics regarding the extent to which gender-inclusive language would impact German language material if non-inclusive texts were to be re-written. More generally, there is a lack of data on the proportion of text that actually refers to human beings, i.e. that could potentially be subject to gender-inclusive language. There are only a few studies dealing with the quantitative-empirical analysis of personal nouns in written texts from a linguistic perspective, e.g. corpus-based studies targeting linguistic entities that are easy to recognise automatically, such as pronouns, especially in English texts (e.g. Saily, Nevalainen & Siirtola 2011; Zeng 2023; cf. also Motschenbacher 2015: 34–35); some studies are enriched with manual analyses (e.g. Baker 2010; Rosola et al. 2023); there is also an increasing interest in the topic in economic disciplines (e.g., for German texts cf. Eugenidis & Lenz 2022). However, even if personal nouns could be detected automatically in the future (cf. Sökefeld et al. 2023 for an initial attempt), the problem of identifying reference would remain unsolved. This is especially relevant for the so-called masculine generic, which is the main focus of gender-inclusive language, and which cannot be distinguished from masculine specifics by mere form (cf., e.g., Schmitz, Schneider & Esser 2023). Manual annotation is therefore necessary for our research question. Our starting points are studies in which German personal nouns are analysed manually and on the basis of small linguistic datasets (e.g. Doleschal 1992; Kusterle 2011; Pettersson 2011). The annotation system for the present study was developed on this basis (cf. Section 3.1). The article is structured as follows: In Section 2, we provide more background information on gender-inclusive language and personal nouns in German. In Section 3, we describe the method of our study, followed by results and discussion in Section 4. We conclude our paper with a brief outlook towards the future in Section 5.

---

[1] Own translation, original: "Genderdeutsch macht einen Text schwerfällig".
[2] Own translation, original: "Mit diesen Verdoppelungen und Sonderzeichen wird die Sprache zudem für Ausländer, die erwägen, Deutsch zu lernen, noch komplizierter und abschreckender."
[3] Own translation, original: "Verständliche, lesbare und zugängliche Sprache wird durch Gendern nicht gewährleistet. Sternchen und Passivkonstruktionen machen Texte leseunfreundlich und länger."
[4] Own translation, original: "Als Nachteil [des Genderns, inbes. Doppelformen] gilt der sprachliche Aufwand, dessen Wiederholung Texte lang, eintönig und sexusfixiert macht."

## 2. Gender-inclusive language and personal nouns in German

German is a grammatical gender language with three grammatical genders (masculine, feminine, neuter). There is a mix of semantic and formal regularities to assign grammatical gender to words, but, according to Hellinger and Bußmann, for "approximately 90% of German monosyllabic nouns, gender class membership can be predicted from morphophonological criteria" (2003: 143). Gender assignment of personal nouns, however, requires special attention, as it is often driven by lexical-semantic factors: "The assumption that, in principle, the assignment of a German noun to one of the three gender-classes is arbitrary, is unfounded in the field of animate/personal nouns, where explicit relations between grammatical gender and the noun's lexical specification can be formulated" (Hellinger & Bußmann 2003: 146).



According to Hellinger and Bußmann (2003, pp. 150-160), when referring to persons in German, we can distinguish between personal nouns that specify referential gender by grammatical, lexical or morphological means:

1. Specification of referential gender by grammatical means: Singular personal nouns in German that are derived from adjectives (like *gesund*, 'healthy') and verbs (*studierend*, present participle of *studieren*, 'to study'; *abgeordnet*, past participle of *abordne*n, 'to delegate') use the grammatical gender of the articles (e.g. *die* (f.) *Gesunde* vs. *der* (m.) *Gesunde*) or the adjective inflection (e.g. *eine Abgeordnete* (f.) vs. *ein Abgeordenter* (m.) "to make referential gender explicit or overt" (also called 'Differentialgenus'/'double gender"; cf. Hellinger & Bußmann 2003 p. 150). Gender specification in these nouns is neutralized in the plural, since articles and other determiners do not vary for grammatical gender in the plural (*die Gesunden* (m./f.pl.), *die Studierenden* (m./f.pl.), *die Abgeordneten* (m./f.pl.)). Some indefinite pronouns can also have this kind of double gender, e.g. *keine/jede* (f.) vs. *keiner/jeder* (m.) or *keines/jedes* (n.; 'no'/'each, every'); some others are grammatically invariable and always masculine (so-called generic pronouns like *jemand* or *niemand*, 'somebody', 'nobody').

2. Specification of referential gender by lexical means: Gender-specification by lexical means is often realised in compounds that denote occupations and functions, containing the second elements *-mann* ('-man') or *-frau* ('-woman') (like *Feuerwehrmann, Feuerwehrfrau*, 'firefighter'). Additionally, there are nouns where referential gender is encoded in the lexical meaning and usually results in lexical pairs, e.g. *die Tante* 'aunt' vs. *der Onkel* 'uncle'; *die Tochter* 'daughter' vs. *der Sohn* 'son'. In this category, grammatical gender is congruent with extra-linguistic gender. In the following, we call these nouns lexical gender nouns.

3. Specification of referential gender by morphological means: The most prominent way to specify referential gender in German is to use suffixes that make the noun gender-specific. This function is mostly carried out by the feminising suffix *-in* which can be attached to most masculine derivation bases (e.g. *Arbeiter/Arbeiterin*, 'male/female worker'; *Maler/Malerin*, 'male/female painter') (for more marginal feminising suffixes, cf. Doleschal 1992: 27–29; Hellinger & Bußmann 2003: 152–153; compare the superficially similar, but functionally



different suffix –ess in English Stefanowitsch & Middeke 2023). There is only a small set of feminine bases that are used to derive masculine terms in the human domain: *Braut/Bräutigam* ('bride/bridegroom') *Witwe/Witwer* ('widow/widower') and *Hexe/Hexer* ('witch/witcher').

As an option to neutralize referential gender in German (besides the use of plural forms of nominalized adjectives/participles), we can use collectives (e.g. society, group, family, etc.) and epicene nouns, i.e. lexical items with a fixed grammatical gender that can refer to any extra-linguistic gender. These nouns occur in all three grammatical genders (e.g. *die Person*, f. 'person', *der Mensch*, m. 'human being'; *das Kind*, n. 'child') (Corbett 1991: 67; cf. Klein 2022).

Resulting from these gender differentiations, German has various kinds of pair forms when denoting humans: a) double gender pairs (e.g. *der Kranke/die Kranke*, 'sick person'); b) (asymmetrical) lexical pairs (*Krankenschwester/Krankenpfleger* 'nurse/male nurse'; *Vater/Mutter* 'father/mother'); c) masculine forms with feminine derivations (e.g. *der Arzt/die Ärztin*, 'male/female doctor'). All of these are semantic minimal pairs, i.e. they have the opposing semantic features +male/-female and +female/-male (Diewald 2018: 290–293). Within these pairs, the masculine form usually has two functions: first, as a masculine specific and, second, as a so-called 'masculine generic'. The term denotes the use of the masculine form to refer to a group of people whose gender is unknown, irrelevant, or ignored (like *Wissenschaftler*, 'scientists', for a group of scientists) and is used in many natural and grammatical gender languages to refer to people in a generic way (Hellinger & Bußmann 2001). Parallel to that, there can be feminine generics in German (e.g., referring to all scientists with the term *Wissenschaftlerinnen*), but these are very rare compared to masculine generics and are often used consciously as a means of gender-inclusive language, e.g. in recent years in the newspaper *Die Zeit* (Dülffer 2018).

Whether a personal noun refers specifically to a male person or generically to a group of unknown gender cannot be decided based on the surface form. On the one hand, the masculine and feminine forms of a personal noun like *Wissenschaftler* ('scientist') may be used as semantic minimal pairs to



refer to male vs. female individuals. Consider the following context: *Das Podium bestand aus drei Wissenschaftlern und einer Wissenschaftlerin* ('the panel consisted of three male scientists and one female scientist'). In this case, the linguistic category grammatical gender reflects the referential gender of the extra-linguistic referents (i.e. the masculine form maps onto male referents, the feminine form maps onto female referent). On the other hand, the grammatically masculine terms are also used to refer to mixed groups of people, to people of unknown gender, or in contexts where gender is presumably irrelevant, e.g. in contexts like die *Wissenschaftler sind sich bislang nicht einig* ('the scientists [m.pl.] do not yet agree'). Here, grammatical gender is assumed to be a neutral category, i.e. not carrying information about referential gender. The reference of the superficially identical lexemes is only resolved in context, which is why the question of whether a masculine form is used specifically (i.e. to designate individual male referents) or generically (i.e. for indefinite referents or mixed groups) cannot yet be detected automatically (Sökefeld et al. 2023: 38) and must be examined individually for each case (Elmiger, Schaefer-Lacroix & Tunger 2017: 64).

The use of masculine generics to denote all genders is subject to controversial societal and academic debates (Müller-Spitzer 2022a; Müller-Spitzer 2022b; Pusch 1984; Simon 2022; Trutkowski & Weiß 2023). Proponents of gender-inclusive language usually do not accept it as a gender-neutral way to indicate person reference (e.g. Acke 2019: 308; Hellinger & Bußmann 2003: 160–161). Opponents of new forms of gender-inclusive language, by contrast, consider the masculine generic to be gender-neutral 'by default' (sometimes based on Becker's assumption of conversational implicatures, cf. Becker 2008; or based on Jakobson's concept of markedness, cf. Eisenberg 2020; Meineke 2023; or on historical data, cf. Trutkowski & Weiß 2023). However, many psycho- and neurolinguistic studies find that the so-called masculine generic is not always understood neutrally but rather activates a male bias (Glim et al. 2023; e.g., Gygax et al. 2008; Körner et al. 2022; Zacharski & Ferstl 2023), i.e. "does not represent men and women equally well" (Glim et al. 2023: 2). These effects are, at least in part, due to the grammatical properties of German, in which the masculine form fulfils the double-function outlined above (Garnham et al. 2012). In addition, gender stereotypes and true gender ratios in the respective groups modulate these effects (Gygax, Garnham & Doehren 2016).



In 2018, the German Personal Status Act was amended to introduce a third gender option (called *divers*) for intersexual individuals. These developments have made the question of how to best address people beyond the binary spectrum more urgent (for research on this topic in other languages cf., e.g., Decock et al. 2023; Kaplan 2022; Thorne et al. 2023). An option already well established in the language system is to use neutralisations such as epicene nouns or derivatives of adjectives and verbs in the plural. However, besides established feminization strategies (pair forms like *Lehrerinnen und Lehrer*, 'female and male teachers'), so-called gender symbols came into use. They are intended to encompass all gender identities (e.g. Lehrer*innen, Lehrer:innen, 'teachers of all genders'; cf. Friedrich et al. 2021; Körner et al. 2022), which a recent psycholinguistic study suggests to be actually the case (Zacharski & Ferstl 2023).[5] The symbols work particularly well in the plural because dependent elements in the noun phrase remain the same for both genders and because the morphological combination of masculine base and feminine derivation suffix is easy to split up with a symbol in the plural. Some qualitative studies have already found tendencies for fewer masculine generics and more gender-inclusive forms (cf. Adler & Plewnia 2019; Elmiger, Schaefer-Lacroix & Tunger 2017; Krome 2020). Quantitative studies on the use of these symbols are still scarce (e.g. Sökefeld 2021; Waldendorf 2023).

In the wake of this debate, many public bodies, large companies and other institutions are now issuing guidelines on gender-inclusive language (cf. links to guidelines of German-speaking cities; Müller-Spitzer & Ochs 2023: 5). However, this new awareness of gender-inclusive language has been accompanied by strong counter-movements that continue to fuel the debate and challenge the ideas behind gender-inclusive language in general (for discussions about gender-inclusive Spanish, cf. Banegas & López 2021). Opponents often argue that the gender symbols are not part of the German language/spelling system and thus should be regarded as 'mistakes' (e.g. Eisenberg 2022). It is also claimed that they distract from the essential content of a text, or that they make texts harder to read, especially for children, L2 learners, or the visually impaired (e.g. Kalverkämper 1979; Münch 2023; Rothmund & Christmann 2002). However, we argue that such effects are only likely if gender-inclusive texts are very different from those that are not gender-inclusive. This is the point of departure for our main research question. By analysing, on the basis of a large corpus, how much of a text would potentially be affected by gender-inclusive language in German, we are able to contribute quantitative data to assess the actual relevance (measured as the proportion of affected textual material) of these claimed effects.

---

[5] Schunack and Binanzer give an extensive overview of possible forms (2022: 4), and new proposals to integrate the new forms into the German grammatical system are discussed by Völkening (2022).

# 3. Method

## 3.1    Corpus and source selection

Our study is based on the German Reference Corpus (DeReKo; Kupietz et al. 2010; Kupietz et al. 2018), from which a sample of texts was selected (cf. Section 4.2). These were taken from four sources: DPA (Deutsche Presseagentur 'German Press Agency') and the magazines Brigitte, Zeit Wissen, and Psychologie Heute. The DPA texts are the central resource for the study. There are several reasons for this. First, DPA is the biggest news agency in Germany, and its reports are distributed to almost all major radio stations and daily newspapers (Pürer & Raabe 2007: 29, 327). Its texts are often re-printed verbatim or only with slight variations. Second, DPA is obliged to be impartial and independent from political parties, worldviews, economic and financial groups, and governments,[6] meaning that its reports can be considered as objective as possible. Third, DPA only recently announced its decision to use gender-neutral language from now on,[7] meaning that DPA texts from before 2021 are not already gender-inclusive and thus serve as a good basis to investigate non-gender-inclusive language. Therefore, we only included texts from the years 2006-2020 to tackle our research questions. Fourth, our aim was to annotate whole texts, as selecting only excerpts could have undesirable biasing effects, e.g. a masculine form could be interpreted as generic, although earlier/later in the text a specific referent is introduced. DPA releases have an average length of 339 tokens in DeReKo (cf. Table 1) and are therefore relatively short, making them well suited for whole-

---





text analyses. To check whether similar patterns would be found in entirely different media outlets, we created a control corpus containing longer texts. The magazines Brigitte, Zeit Wissen, and Psychologie Heute were selected because they have a more general societal outlook and/or cover popular science topics. All three are issued by different publishers, minimizing the influence of publisher-specific guidelines.[8]

## 3.2  Sampling

In the overall corpus (DeReKo), there are 2,322,095 documents available for all four sources. The sampling process was based on the number of words (tokens) per document. For each source, we calculated the 5th and 95th percentile of token counts. For DPA, the interval is [I = 87, 837], i.e. 90 % of all DPA documents are between 87 and 837 words long and were selected for the sampling procedure. The values for the other sources, as well as median (50th percentile) and mean values are given in Table 1. The upper bound for the journal sources is generally higher than for the DPA documents, i.e. there are more longer documents in the magazine sources. This is also reflected in the median and mean token counts for the four sources.

| | | DPA | Brigitte | Psychologie Heute | Zeit Wissen |
|---|---|---|---|---|---|
| Available in DeReKo | Number of Texts | 2,298,618 | 17,055 | 4,130 | 2,292 |
| Sampling Criteria | 5th percentile (lower bound) | 87 | 53 | 59 | 101 |
| | median | 262 | 411 | 359 | 514 |
| | mean | 339 | 700 | 676 | 1,052 |
| | (standard deviation) | 251 | 852 | 848 | 1,187 |
| | 95th percentile (upper bound) | 837 | 2,478 | 2,751 | 3,525 |
| Available for Sampling | Number of Texts | 2,071,006 | 15,387 | 3,721 | 2,065 |
| Annotated Sample (annotated by both annotators, | Number of Texts | 184 | 35 | 36 | 6 |
| | median | 223 | 332 | 415 | 626 |
| | mean | 266 | 499 | 643 | 663 |

---

[8] *Brigitte*, published by Gruner+Jahr, is a women's magazine covering a wide range of social issues (approx. 241,000 copies sold, biweekly: https://en.wikipedia.org/wiki/Brigitte_(magazine) [last accessed: 30 January 2024]). *Psychologie Heute* is a popular science magazine on psychology, published monthly by the Beltz publishing group (approx. 63,000 copies sold: https://de.wikipedia.org/wiki/Psychologie_Heute [last accessed: 19 December]), and *Zeit Wissen* is a popular science magazine published by the Zeitverlag (approx. 97,000 copies sold, bimonthly: https://en.wikipedia.org/wiki/Zeit_Wissen [last accessed: 30 January 2024]).



| | | | | | |
|---|---|---|---|---|---|
| without punctuation) | standard deviation | 154 | 471 | 682 | 452 |

Table 1: 5th percentile, median value (50th percentile), mean value and 95th percentile for word (token) counts in the four sources. Where applicable, figures are rounded to the nearest integer.

We randomly sampled a fixed number of documents that fall into the inner 90 % of token counts (between the 5th and 95th percentile). For DPA, we sampled 190 documents, and for the journal sources 40 documents each, i.e. we had a total of 310 sampled documents. After annotation, 261 texts remain in the corpus (for details, cf. Section 3.3). Their token counts are summarized in Table 1 under 'Annotated Sample'.

## 3.3    Annotation process

The aim of the manual annotation conducted for this study was to find all tokens that would have to be changed if the text was reformulated in a gender-inclusive way. Our annotations focus on expressions that refer to natural persons, usually heads of noun phrases (NPs) in the form of nouns or pronouns (cf. Stede 2016: 55). Accordingly, we follow an action-theoretical conception of reference based on the interpretation of the target item in the given context (Pettersson 2011: 57). In addition to the head of the noun phrase, dependent elements in the NPs are annotated. For that, we decided to apply a strict bottom-up approach, i.e. to identify the head first and then select the elements that are dependent on it (especially articles and attributive adjectives, cf. Table 2, as these can theoretically be affected by gender-inclusive language, as opposed to genitive constructions or prepositional phrases). The manual that served as the basis for the annotations was developed over the course of several months. Modifications were implemented after each training round, when we could see difficulties and uncertainties regarding the application of the manual. Building upon the insights from these pre-tests, the annotation scheme underwent refinement and expansion. The more elaborate annotation scheme was then used in its final version for the present study, which was conducted from December 2022 to March 2023. Two student assistants (in the following called annotators A and B) annotated the texts simultaneously. 261 of the 310 sampled documents were annotated by both annotators,[9]

---

[9] Of the 310 sampled texts, 261 were annotated by both annotators. 34 were annotated only by annotator A because annotator B left our institution before being able to complete their annotations (1 *DPA*, 33 *Zeit Wissen*). Unfortunately, 15 texts were not annotated at all (5 *DPA*, 5 *Brigitte*, 1 *Zeit Wissen*, 4 *Psychologie Heute*) because of user errors within the annotation tool. We considered re-annotating the missing texts but dismissed the idea because we did not want to introduce potential biases from a third annotator. We



yielding an overall inter-annotator agreement of 77.89%. The version of the annotation scheme used for this study consists of eleven categories with various sublayers. The decision tree in Figure 1 illustrates the dependencies between them. Further information about the decision tree, the layers, the annotation procedure, and the inter-annotator agreement can be found in the supplementary material.[10]

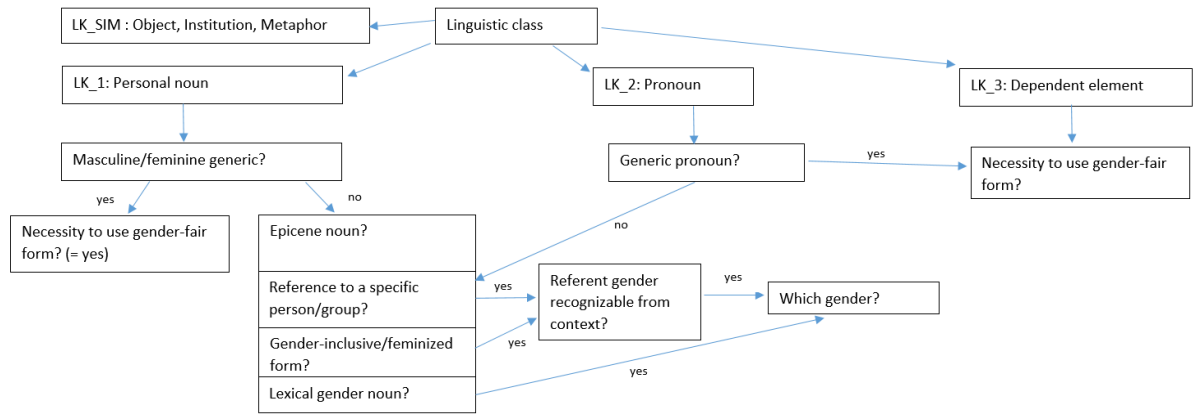

Figure 1: Decision tree for the annotation software and process.

As 'necessity to use a gender-inclusive form' is the central category for our study, it is described in more detail here rather than exclusively in the supplementary material. Within the annotation procedure, it is necessary to indicate for each annotated token whether it would need to be replaced by another form in order to make the text gender-inclusive. For personal nouns, this is usually the case if the form is annotated as a masculine or feminine generic. Regarding pronouns, this is only the case for generically referring personal pronouns.[11] Dependent elements need to be adjusted only if the head of the NP would be subject to change – however, dependent elements need to be thoroughly checked to determine whether they would actually change form in case of an adjustment to gender-inclusive language (e.g. most attributive adjectives are identical for both genders: *der kranke Patient*/*die kranke Patientin*; 'the sick [male/female] patient'). Example (1) illustrates this further:





a. *Während die einst potenten Sozialdemokraten im Bund unter ihrem Parteichef Kurt Beck in der Krise stecken, träumen die traditionell schwachen bayerischen Genossen von der Machtübernahme im Freistaat.* ('While the once-powerful Social Democrats at Federal level are in crisis under their leader Kurt Beck, their traditionally weak Bavarian comrades dream of taking power in the Free State [Bavaria].' (DPA08_JUL03207)

In this sentence, only two nouns are annotated as having a 'necessity to use gender-inclusive form' (printed in bold). The dependent elements in the noun phrase would not need to be changed, as can be seen in Table 2: Even if the heads were changed to gender-inclusive forms, the dependent elements would remain the same. The excerpt also contains a specific male person (*Kurt Beck*) and a masculine role description, *Parteichef* ('party leader'), referring to him. Accordingly, *Parteichef* is annotated as a personal noun with specific male reference.

| 1 | die einst potenten | *Sozialdemokraten* | […] | die traditionell schwachen bayerischen | *Genossen* |
|---|---|---|---|---|---|
| 2 | die einst potenten | Sozialdemokratinnen und Sozialdemokraten | | die traditionell schwachen bayerischen | Genossinnen und Genossen |
| 3 | die einst potenten | Sozialdemokrat*innen | | die traditionell schwachen bayerischen | Genoss*innen |

Table 2: Illustration of the necessity to use a gender-inclusive form (1: original sentence, in italics the token that has to be changed in case of using gender-inclusive language; 2: possible reformulation using pair forms; 3: possible reformulation using gender asterisk).

In what follows, we will report analyses based on the 261 documents that were annotated by both annotators.[12] Figure 2 gives an overview of the token count distributions of candidate texts (i.e. all texts after selecting the inner 90 % of token counts for each source) and the 261 texts on which we base our analyses. Comparing these distributions, we can conclude that the texts reported here provide a good reflection of the underlying token count distributions for DPA, Brigitte, Psychologie Heute, and Zeit Wissen.[13]

---

[12] Information on data and code availability can be found in Section 5 of the Supplementary Material.

[13] Note that only six texts from *Zeit Wissen* remain in the final analysis. It is particularly noticeable that no longer text from the upper end of the distribution for *Zeit Wissen* is included. However, we will conduct no general analysis of single sources from the control corpus. Rather, the remaining six *Zeit Wissen* texts enter the larger collection of texts together with *Brigitte* and *Psychologie Heute* to serve as a control corpus with which results from DPA can be compared.



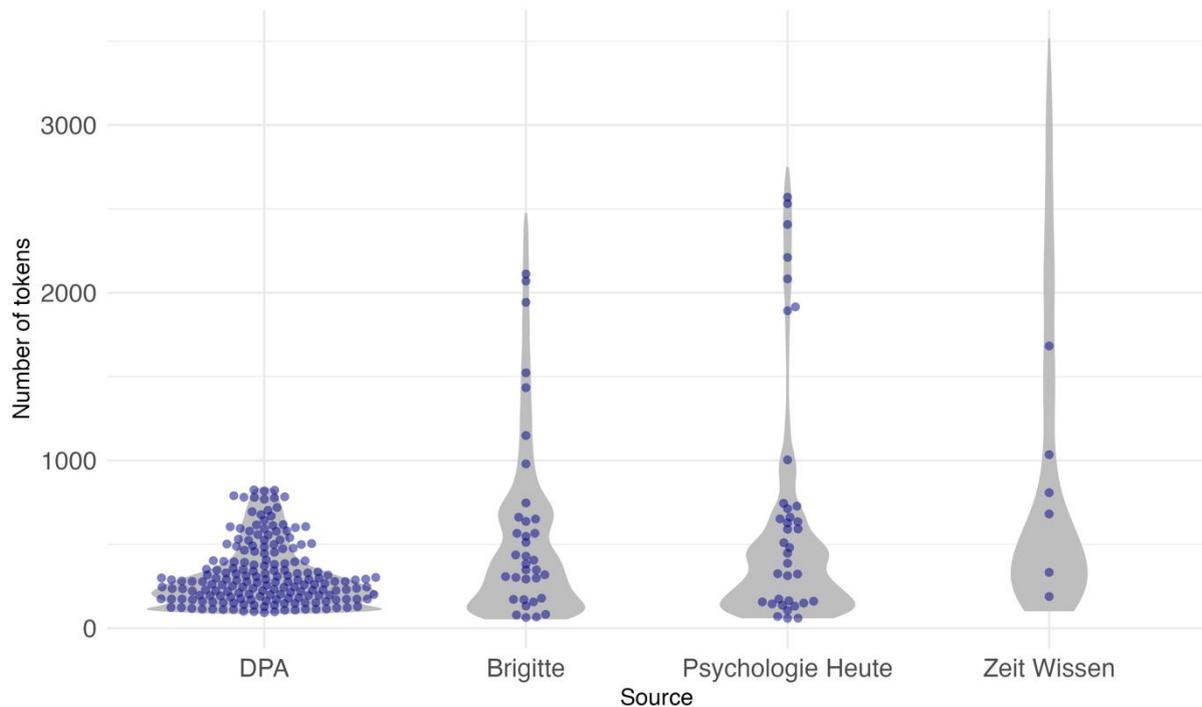

Figure 2: Distribution of token counts (y-axis) for the candidate texts (grey violins), i.e. all texts with a token count in the inner 90% of all texts from this source (x-axis). Data points represent token counts for all 261 texts reported in the remainder of the paper (one point per document).

# 4. Results & Discussion

## 4.1    Person reference and linguistic classes

In total, the 261 texts annotated by both annotators comprise 120,626 tokens.[14] Without punctuation, 93,533 tokens remain.[15] Of these, 11,375 tokens (12.2%) were annotated by at least one annotator as having person reference (i.e., as belonging to the linguistic classes 1-3: personal noun, pronoun, dependent element). The annotators agreed on the linguistic class of 8,840 (A = B; 77.71%) of these tokens; another 675 (5.93%) were annotated by both, but with diverging linguistic classes (A ≠ B); 1,860 tokens (16.35%) were annotated by only one annotator (A v B). This means that the vagueness regarding which token can be considered person reference is roughly 16%. Importantly, this vagueness (or insecurity) is distributed unequally across linguistic classes. Dependent elements (LK_3) caused the most insecurities, constituting 58.06% of all tokens that were annotated only once. From what we discussed earlier regarding phrase structure, it is probable that this is due to uncertainties about which elements belong to an NP and therefore have person reference. Hence, an important takeaway for future studies is the necessity to enhance the training of student assistants in the domain of phrase-structure grammar. With 27.37%, personal nouns (LK_1) range second regarding non-matching annotations. Pronouns were the least problematic, making up 13.82% of vagueness. The remaining proportion of insecurity is attributed to nouns that superficially look like personal nouns but actually refer to objects or institutions or are used metaphorically (e.g. *Partner* to refer to a country). For this study, we only consider the 8,840 tokens with matching annotations to represent reliably the amount of person reference in the annotated texts. Personal nouns are the biggest category here with 3,196 tokens (3.42% of all tokens; Sökefeld et al. 2023 find a similar proportion of personal nouns in their automatic detection tests), followed by dependent elements (3,097 tokens; 3.31%) and pronouns (2,547 tokens; 2.72%).

---

[14] The 34 texts that were only annotated once (by annotator A) comprise 30,764 tokens (with punctuation). These are excluded from the present study as we want to focus on those tokens that have matching annotations from both annotators.

[15] We decided to exclude punctuation from the total token count because including them would have led to an under-estimation of the share of tokens that would be affected by the use of gender-inclusive language.

All measures reported so far refer to all documents in the corpus as one large list of tokens. However, in order to accurately assess the relevant proportions of tokens, we have to consider the document level (which was also our level of sampling). We therefore determined the proportions for each document and then calculated overall means, weighted by the number of tokens each document contributes to this mean. For each value of the weighted mean, we report 95% confidence intervals according to a hypergeometric distribution in brackets. This is the appropriate method in this case because the annotated texts were sampled from the candidate texts without replacement.[16] Figure 3 shows the proportions of tokens with person reference for the DPA and the control (i.e. *Brigitte*, *Psychologie Heute*, *Zeit Wissen*) corpora. While the mean for all DPA documents is 7.99% (7.75% – 8.23%), it is significantly higher for the control corpus at 11.06% (10.77% – 11.35%). The mean of all sources taken together is 9.45% (9.26% – 9.64%).

---

[16] In Section 4 of the supplementary material, we also provide unweighted mean proportions with bootstrapped 95% confidence intervals.



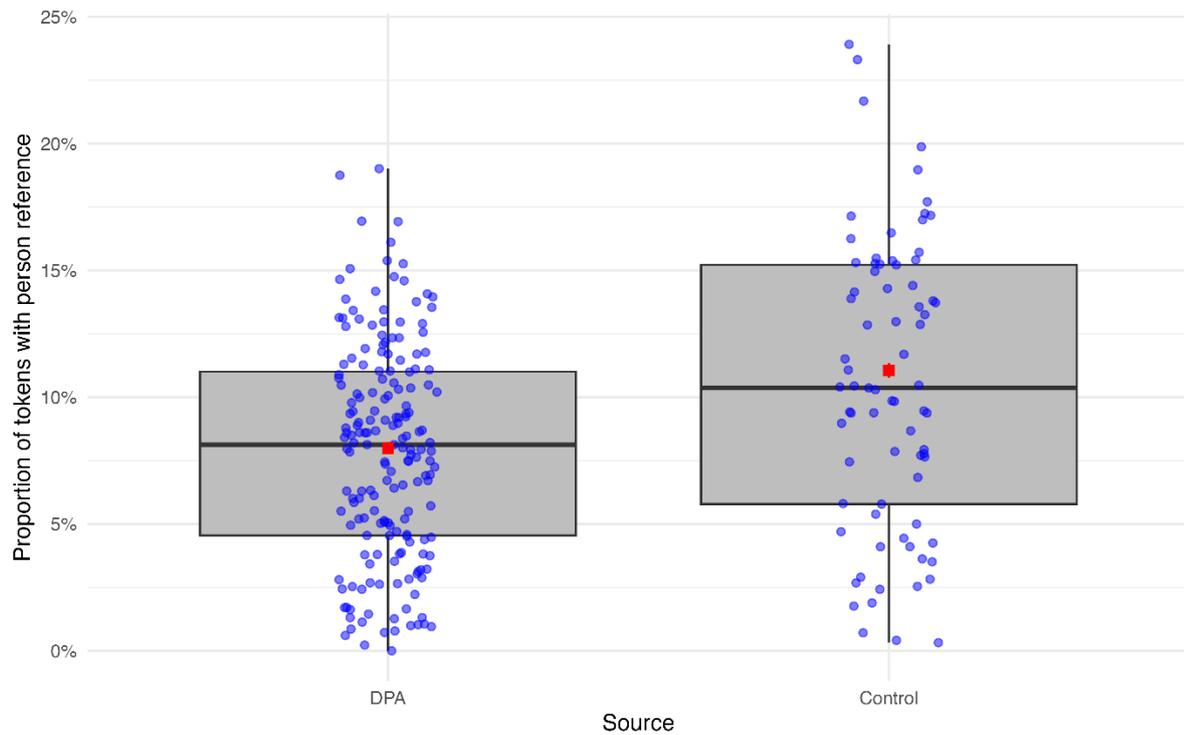

Figure 3: Proportion of tokens with person reference in DPA and control corpus. Data points represent documents; the red square indicates the mean, including error bars that symbolize the 95% confidence intervals (sometimes these are fully covered by the square – e.g. in the left boxplot).

## 4.2  Necessity to use gender-inclusive language

The annotation category 'necessity to use gender-inclusive form' holds a pivotal role in addressing the primary research question of this study: determining the extent to which tokens within press texts would undergo changes due to the adoption of gender-inclusive language. For DPA, the average share of tokens that would be affected by gender-inclusive re-editings is 0.73% (0.66% – 0.81%), whereas it is 1.18% (1.09% – 1.29%), and therefore significantly higher, for the control corpus (cf. Figure 4). If we take all sources together, we get a proportion of 0.95% (0.89% – 1.01%) that would be affected by gender-inclusive language. Considering only tokens with person reference, an average of 9.13% (8.25% – 10.08%) would be affected by gender-inclusive language in DPA and 10.67% (9.82% – 11.56%) in the control corpus. Here, the difference between the corpora is not significant. Still, it indicated that, while a lower share of all DPA tokens would be affected by gender-inclusive language, the proportion is slightly higher than in the control corpus when looking only at person references. This is especially interesting because DPA has a significantly lower share of person references than the control corpus, meaning that, even though fewer person references are encountered in DPA, more of these would be affected by gender-inclusive language. Taking all



sources together, an average of 9.99% (9.37% − 10.63%) of person references would be subject to gender-inclusive language. We can therefore record three central measures so far: all sources taken together, on average a) 9.45% of tokens are (parts of) person references; b) 0.95% of all tokens would be affected by gender-inclusive language; and c) 9.99% of all tokens with person reference would be affected by gender-inclusive language.[17]

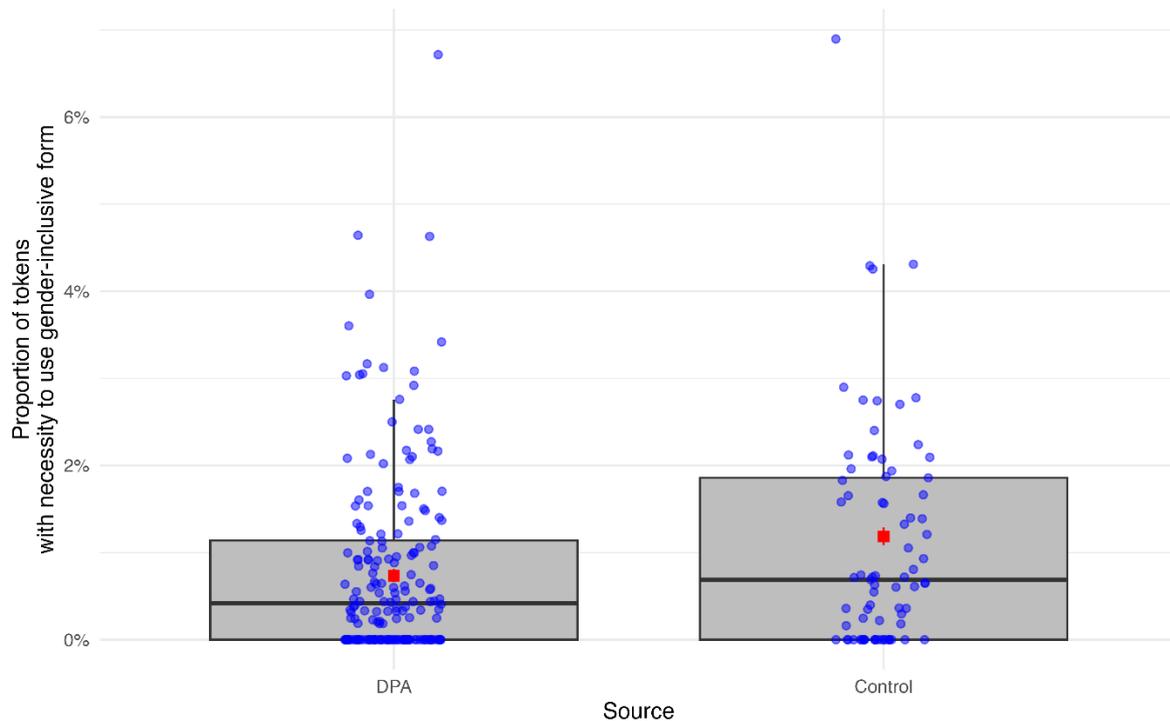

Figure 4: Proportion of tokens with necessity to use gender-inclusive language in DPA and the control corpus.

Figure 5 shows that the largest proportion of affected tokens (tokens that would have to be changed) belongs to the category 'personal nouns' (799 tokens overall; 90.08% of the total of 887), i.e. gender-inclusive language would mostly affect nouns. All affected personal nouns are masculine generics, highlighting that they are the focus of gender-inclusive language. The average proportion of personal nouns that would be changed by gender-inclusive language across all the documents is 25.00% (23.51% − 26.54%). There are 6 documents in which all personal nouns would be subject to change (i.e. the dots at the 100% margin), but far more documents in which none of the personal nouns would need to be changed (N = 81, dots at the 0% margin). More details on the amount of documents that would be affected by changes can be found in Section 4.3. For the other two linguistic classes, the proportion is only marginal – in most documents, none of the pronouns or dependent elements would be subject to change. This is especially true for pronouns, where the average proportion is 0.12% (0.02% – 0.34%). For dependent elements, it is 2.62% (2.08% – 3.24%), with one outlier document in which about 66% of dependent elements would be changed if gender-inclusive language were used in the document. Gender-inclusive re-editings would therefore rarely interfere with the grammar of the extended noun phrase.

---

[17] We suppose that the perceived 'omnipresence' of gender-inclusive language might stem from a conflation of these numbers in lay perspectives. As an opener to a conference talk, we asked: *What percentage of tokens in press texts would have to be changed (from non-gender-inclusive to gender-inclusive)*? Most of the audience thought '10%' was the right answer (the options were: 1%, 3%, 7%, 10%). It could be the case here that people correctly guessed the amount of *person reference* in texts, and then assumed that *all* of these tokens would need to be changed; or they correctly assumed the share of *person references* that would be affected instead of the share of *all* tokens. Our analyses show that gender-inclusive language would leave the biggest share of tokens *and* person references unchanged.

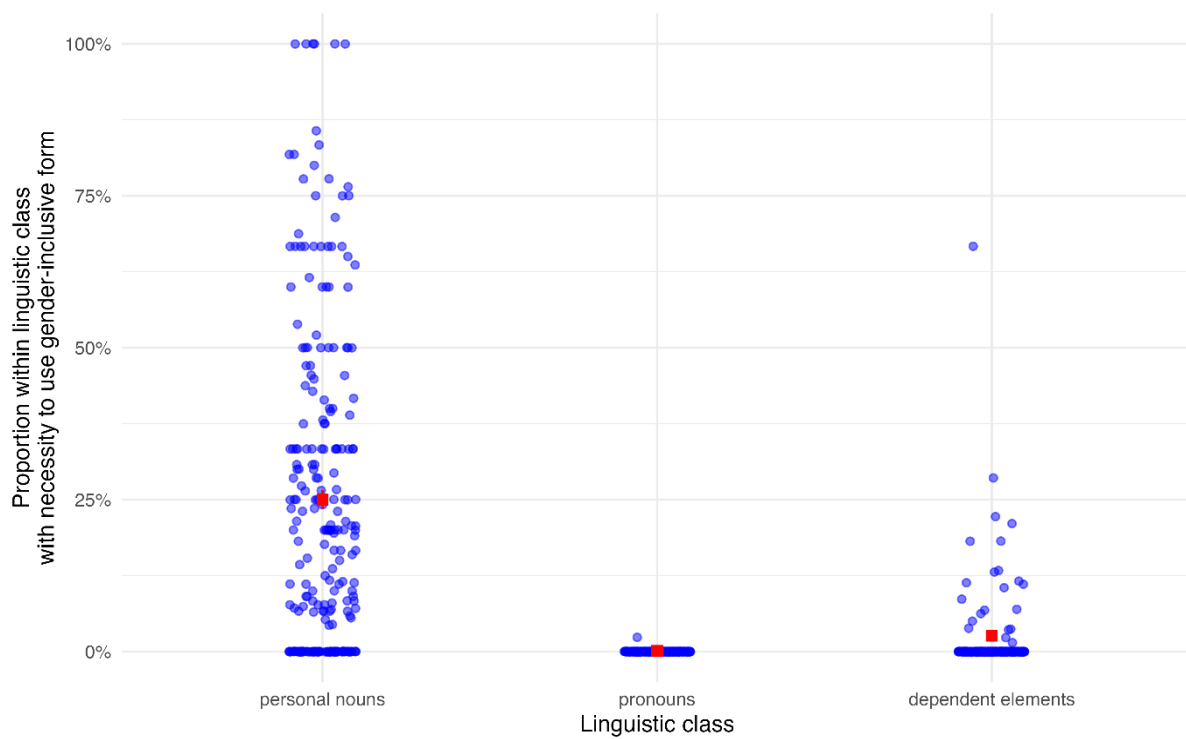

Figure 5: Necessity to use gender-inclusive form split by linguistic classes (all documents taken together).

## 4.3   Personal nouns

Because personal nouns are a) the most frequent linguistic class, b) the class with the highest proportion of tokens that would be affected by gender-inclusive language, c) the linguistic class with the most diverse annotation layers (cf. Supplementary Material, Section 1), and d) the linguistic class most relevant to the study of the linguistic representation of people in texts (cf. e.g. Hellinger & Bußmann 2003: 143), the category is discussed in more detail here.

First, we report the distribution of annotation layers for personal nouns. Taking all sources together, we see a prominence of epicene nouns (27.32%, 25.78% – 28.90%), closely followed by masculine



generics (24.97%, 23.48% − 26.51%), and masculine specifics (22.93%, 21.49% − 24.43%). Lexical

gender nouns (9.95%, 8.93% − 11.04%) and feminized forms (8.04%, 7.12% − 9.04%) are

significantly rarer. We find no nouns that were annotated as feminine generics by both annotators.

Figure 6 shows that the distribution of layers for personal nouns varies considerably between the two

corpora. In DPA, masculine specifics are by far dominant (mean share of 36.47%, 34.10% −

38.89%), especially compared to the control corpus, where this category only amounts to an average

share of 9.48% (8.09% − 11.02%). For all other categories, it is the other way around. The average

shares of epicenes, masculine generics, lexical gender nouns, and feminized forms are always higher

in the control corpus. We can deduce that DPA predominantly reports on specific male persons, using

masculine forms, whereas the other sources tend to report more unspecifically, i.e. making use of

gender-neutral forms (epicenes) and masculine generics. Referent gender is mostly specified by

lexical gender nouns in the control corpus (e.g. Frau, Mann 'woman, man'), while these forms are

infrequent in DPA.

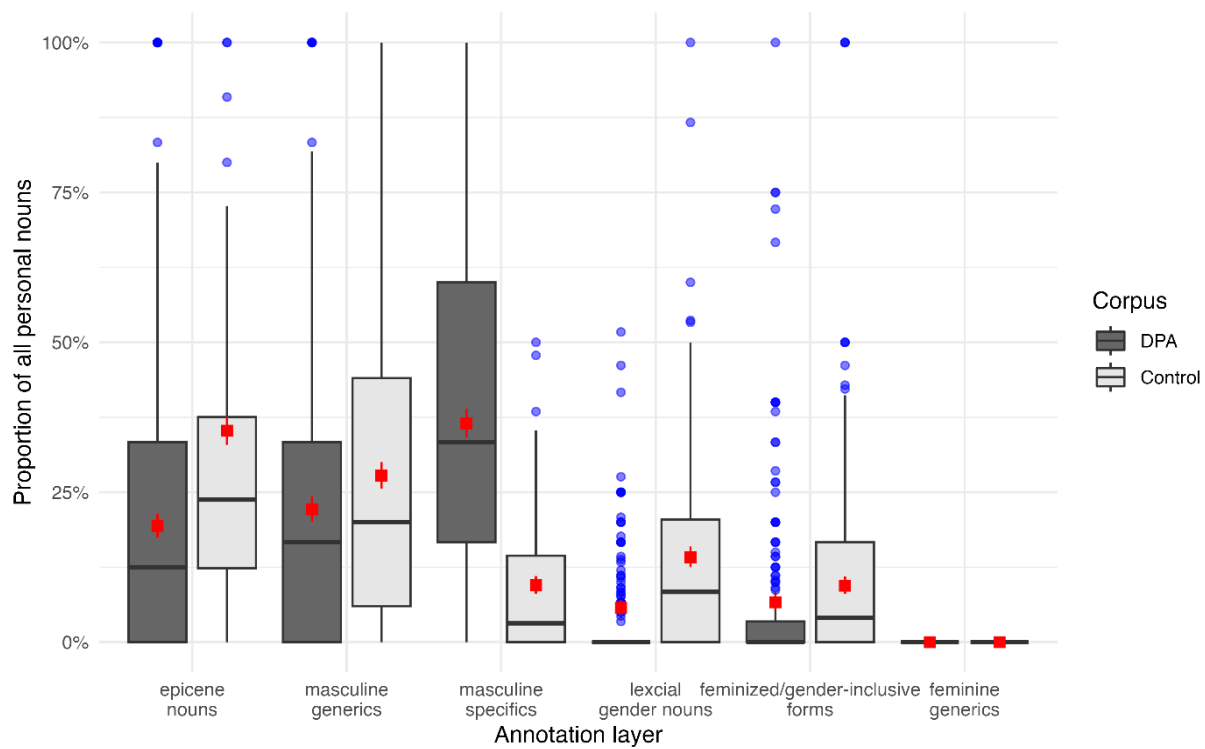

Figure 6: Types of personal nouns by corpus. Only outlier documents are shown as data points.

This is closely related to the differences in gender distributions of individuals reported on in the

sources (cf. Figure 7). In DPA, there is a clear male bias. If referent gender is recognizable from



context, a mean share of 80.37% (77.46% − 83.05%) of these tokens refer to men. Only an average of

19.01% (16.37% − 21.89%) refer to women. This strong male dominance in news reporting is in line

with findings in other studies (e.g., Lansdall-Welfare et al. 2017; Saily, Nevalainen & Siirtola 2011).

In the control corpus, however, the bias disappears: the average share of tokens referring to women

(52.87%, 48.56% − 57.14%) is even higher than for men (45.29%, 41.04% − 49.59%). Brigitte has

the biggest influence here – a mean of 60.54% (54.32% − 66.51%) of tokens for which referent

gender is identifiable refer to women, while only 38.70% (32.76% − 44.90%) refer to men. No non-

binary referents were identified. 'Group' reference (i.e. to mixed groups of men and women) is rare

(N = 11 in all documents taken together) and not discussed further here. These findings indicate

substantial differences in the way different sources portray men and women, which is most likely due

to differences in topics and audiences (cf. e.g., Müller-Spitzer & Rüdiger 2022). However,

comparable corpus studies are needed to draw such conclusions on a reliable basis.

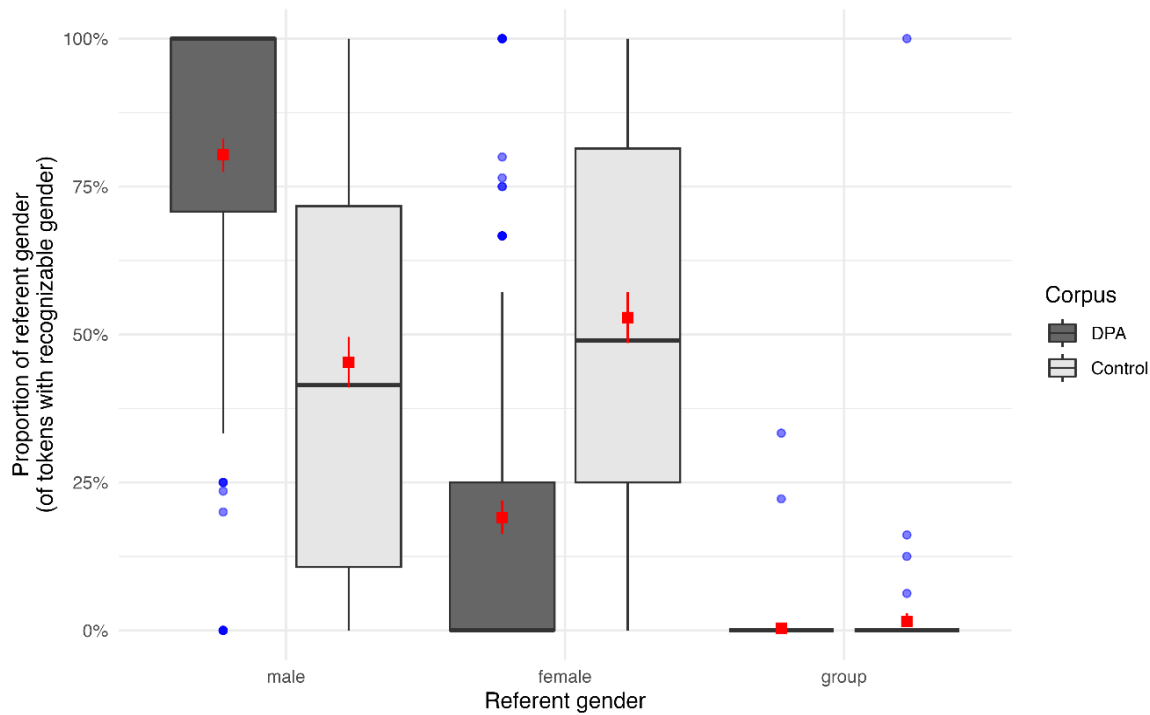

Figure 7: Share of tokens for which 'male/female gender' or 'group' is deducible from context (total amount: tokens for which referent gender is recognizable from context). Only outlier documents are shown as data points.

Our methodology can thus also be used to quantify the occurrences of men and women mentioned in

press texts. It encompasses all personal nouns and therefore goes beyond the analysis of proper

names, which are used by Eugenidis & Lenz (2022), for example, to quantify the proportion of men



and women on company websites. This is especially relevant as media outlets increasingly seek to scrutinize gender proportions within their articles. One example is the renowned German weekly magazine *Der Spiegel*, which conducted an analysis of gender proportions in their texts (Pauly 2021). However, they pointed out that they could not include personal nouns in their evaluations because they used procedures for named entity recognition (Pauly 2021). Our approach could effectively supplement such automated procedures in grammatical gender languages. Additionally, our annotated dataset could serve as training material for developing automatic processes to detect personal nouns, particularly in terms of distinguishing between generic and specific references. The need to supplement automated processes with in-depth annotations is also highlighted by Sökefeld et al. (2023: 38).

Furthermore, our annotations allow us to analyse the distribution of masculine specifics and masculine generics in more detail and to investigate the embedding of masculine generics in actual language use. We can only give a brief insight into these aspects here, focussing on the document level of our data. In total, 116 of the 261 annotated texts (44.44%) contain both masculine generics and specifics. In 55 of these (47.41%), specifics are more common than generics; in 48 documents (41.38%), it is the other way around. Another 22 texts (18.97%) have equal amounts of masculine specifics and generics. In 104 texts (39.85%), we find only one of the forms: 57 (54.81%) have only masculine specifics; 47 contain only masculine generics (45.19%). This means that there are 41 texts (15.71%) in which neither a specific nor a generic masculine is used. In sum, texts with both specifics and generics are most common, followed by texts with only masculine specifics and texts with only masculine generics. Texts without any of these forms are least common. Gender-inclusive re-editings would in sum affect 163 of the 261 annotated documents (62.45%). To put it differently, in more than a third of the documents, nothing would need to be changed if gender-inclusive language was applied.

The prototypical use of the masculine generic is considered to be found in abstract contexts (cf. Zifonun 2018: 49–50) in which no specific individuals are referred to and in which the semantic



category 'gender' is presumably neutralised (ex. b). However, our data show that masculine generics can be used in a diverse set of contextual embeddings and with different levels of referentiality. We find, for example, four cases in which a masculine plural refers to a pair consisting of a man and a woman. They are introduced with their names in the text and then collectively referred to with a masculine generic (ex. c). We also find one masculine form with female reference (ex. 4), which is surrounded by feminized forms and a lexical gender noun that refer to the same person. In many other cases, the masculine generic is used in contexts where a small and specific number of referents (ex. d, e) are introduced but whose genders are not specified in the rest of the text. While having the power to level out the importance of gender in such contexts (Zifonun 2018: 50–51), the masculine generic can also be understood to veil referent genders and make women (and other genders) invisible or at least harder to include cognitively (as is suggested by various psycholinguistic studies on the male bias, e.g. Gygax et al. 2008; Körner et al. 2022; Zacharski & Ferstl 2023). The referential ambiguity of the masculine can be a challenge for recipients, leading to the question whether this challenge is greater than the decoding of gender-inclusive forms (which at least are referentially unambiguous in the sense that they never only refer to men).

> b. *Die Preisträger genießen an Schulen besonderes Ansehen.* ('(The) Award winners enjoy a special reputation at the school.'*)* (from Zifonun 2018: 50)
>
> c. *[...] haben die Psychologen Angela Duckworth und Martin Seligman [...]* ('[...] the psychologists Angela Duckworth and Martin Seligman have [...]') *(PH07_AUG.00032)*
>
> d. *Stylistin und Spielplatzmami, Kinderkutschierer und Großeinkäuferin.* ('Stylist and playground-mummy, children's coachman and bulk buyer.') *(BRG10_JAN.00047)*
>
> e. *Sieben Umweltaktivisten aus verschiedenen Teilen der Welt [...]* ('Seven environmental activists from different parts of the world [...]') *(DPA08_APR.08223)*

As we are publishing the annotated dataset together with this paper (cf. Supplementary Material, Section 5), these different forms of embedding can be further analysed and classified not only by us, but also by other interested researchers.



# 5. Conclusion & Outlook

Research into the connection of gender and language is tightly knitted to social debates on gender equality and non-discriminatory language use. By now, there is a growing body of studies investigating linguistic dimensions of the category 'gender'. Psycholinguistic scholars have made significant contributions, particularly in addressing the male bias associated with masculine generics. However, there exists a demand for more corpus-based studies that investigate these matters within the context of real language usage. In our study, we addressed the question of how much textual material would actually have to be changed if non-gender-inclusive texts were rewritten to be gender-inclusive. In total, one third of all documents we analysed would remain unchanged. Furthermore, we extracted three central values from our data: an average of a) 9.45% of all tokens are (or are parts of) person references; b) 0.95% of all tokens would be affected by gender-inclusive language; c) 9.99% of tokens with person reference would be affected by gender-inclusive language. The small proportion in b) calls into question whether gender-inclusive German presents a substantial barrier to understanding and learning the language, particularly when we take into account the potential complexities of interpreting masculine generics. Besides that, gender-inclusive language would almost exclusively concern nouns, for which there are already numerous strategies for implementing unobtrusive gender-inclusive variants that do not include the disputed gender symbols (e.g. pair forms and epicenes, cf. Steinhauer & Diewald 2017: 118, 132). A recent survey by the German public-broadcasting institution WDR[18] has shown that many of these variants are already widely accepted. As describing and comparing the complexity of linguistic items is a difficult endeavour, our data are mainly intended to provide future research with a quantitative baseline – e.g. to compare the values with proportions of other structures that are considered complex in German

We see especially promising potential in combining our data with automatic extraction procedures for personal nouns (e.g. Sökefeld et al. 2023), e.g. by using our annotations as training data for the

---





recognition of masculine specifics and generics. To further this approach, we are currently in the process of conducting analyses at the lexical level to determine whether certain types of personal nouns (e.g. passive role nouns such as neighbour or citizen, cf. Bühlmann 2002: 174) are more prone to being employed as masculine generics. As corpus-based research into person reference is so far limited to a great degree to German and English, the extension of our approach to more languages would certainly be a fruitful addition to gender and language research.

**Funding Statement**
No funding to declare.

**Data Accessibility**
All the data and code needed to replicate our results, as well as supplementary material, are available here. The current link is temporary as the raw data contains copyrighted material and should only be accessed directly through this link. When our paper is published, the links will be changed - one to a repository where the R script can be downloaded, the other to a tool that provides easy access to both the non-copyrighted and copyrighted corpus material. As this long-term solution allows conclusions to be drawn about the affiliation of the authors, we have implemented this temporary workaround to allow double-blind peer review if necessary.

**Competing Interests**
*We have no competing interests.*

**Authors' Contributions**
Conceptualization: C.M.-S., S.O., S.W., A.K.; Data curation: S.O., S.W., J.-O.R.; Formal Analysis: S.O., S.W.; Investigation: C.M.-S., S.O.; Methodology: C.M.-S., S.O., A.K., J.-O.R., S.W.; Software: J.-O.R.; Visualization: S.O., S.W.; Writing – original draft: C.M.-S., S.O.; Writing – review & editing: C.M.-S., S.O., A.K., J.-O.R., S.W.